\documentclass[conference]{IEEEtran}
\IEEEoverridecommandlockouts
\usepackage{cite}
\usepackage{amsmath,amssymb,amsfonts}
\usepackage{graphicx}
\usepackage{textcomp}
\usepackage{xcolor}
\usepackage{booktabs}

\usepackage{multirow}
\usepackage{algorithm}
\usepackage{algpseudocode}
\usepackage{subfigure}
 \usepackage{subcaption}

\IEEEoverridecommandlockouts
\def\BibTeX{{\rm B\kern-.05em{\sc i\kern-.025em b}\kern-.08em
    T\kern-.1667em\lower.7ex\hbox{E}\kern-.125emX}}
\begin{document}
\IEEEpubid{
  \begin{minipage}{\textwidth}\ \\[36pt]
  Accepted for publication in the 2026 IEEE International Conference on Multimedia and Expo (ICME).\\
  \copyright 2026 IEEE. Personal use of this material is permitted. Permission from IEEE must be obtained for all other uses, in any current or future media, including reprinting/republishing this material for advertising or promotional purposes, creating new collective works, for resale or redistribution to servers or lists, or reuse of any copyrighted component of this work in other works.
  \end{minipage}
}
\title{VA-FastNavi-MARL: Real-Time Robot Control with Multimedia-Driven Meta-Reinforcement Learning}

\author{
    \IEEEauthorblockN{
        \textit{Yang Zhang}\textsuperscript{2}, 
        \textit{Shengxi Jing}\textsuperscript{3}, 
        \textit{Fengxiang Wang}\textsuperscript{3}, 
        \textit{Feng Yuan}\textsuperscript{2}, 
        \textit{Hong Wang}\textsuperscript{1$\dagger$,4}
    }
    \IEEEauthorblockA{
        \textsuperscript{1}School of Artificial Intelligence, Hubei University, Wuhan, 430062, Hubei, China\\  
        \textsuperscript{2}Department of Mechanical and Aerospace Engineering, University of Missouri, Columbia, MO, 65201, USA  \\
        \textsuperscript{3}School of Construction Machinery, Chang’an
University, Xi’an, 710064, Shaanxi, China \\
        \textsuperscript{4}Key Laboratory of Intelligent Sensing System and Security (Ministry of Education), Hubei University, Wuhan, China \\
        \{zhangy1,yfzc8\}@missouri.edu, \{fengxiang.wang,sx.jing\}@chd.edu.cn, hong.wang@hubu.edu.cn
    }
}

\maketitle

\begin{abstract}
Interpreting dynamic, heterogeneous multimedia commands with real-time responsiveness is critical for Human-Robot Interaction. We present VA-FastNavi-MARL, a framework that aligns asynchronous audio-visual inputs into a unified latent representation. By treating diverse instructions as a distribution of navigable goals via Meta-Reinforcement Learning, our method enables rapid adaptation to unseen directives with negligible inference overhead. Unlike approaches bottlenecked by heavy sensory processing, our modality-agnostic stream ensures seamless, low-latency control. Validation on a multi-arm workspace confirms that VA-FastNavi-MARL significantly outperforms baselines in sample efficiency and maintains robust, real-time execution even under noisy multimedia streams.
\end{abstract}


\section{Introduction}
\label{sec:intro}

Recent advances in multimedia technologies have significantly accelerated the development of intelligent systems capable of processing and understanding heterogeneous data from diverse sources. Modern multimedia systems increasingly rely on multimodal learning, where visual, textual, and sensory signals are jointly modeled to enable comprehensive cross-modal understanding and robust representation learning. Representative approaches such as CLIP~\cite{radford2021clip} and Flamingo~\cite{alayrac2022flamingo} demonstrate that aligning information across modalities can substantially improve semantic reasoning and generalization, establishing multimodal learning as a core paradigm in multimedia research~\cite{baltrusaitis2019multimodal}. Beyond representation learning, multimedia research has also expanded toward integrating multimodal perception with sequential decision-making, where agents must interpret and act upon heterogeneous multimedia streams in dynamic environments. For example, approaches such as SayCan~\cite{ahn2022saycan} and RT-2~\cite{brohan2023rt2} incorporate vision-language representations into embodied decision-making, enabling agents to execute high-level semantic instructions, while tasks such as vision-and-language navigation~\cite{wang2019vln} and video-language modeling~\cite{sun2019videobert} further highlight the importance of jointly modeling perception and action in multimedia-driven scenarios. 

To further improve adaptability in such complex environments, multi-agent reinforcement learning (MARL) has been extended with meta-learning mechanisms, allowing agents to rapidly adapt to new tasks and changing conditions. This paradigm, known as Meta Multi-Agent Reinforcement Learning (Meta-MARL~\cite{NEURIPS2023_d1b1a091}), enhances flexibility by bridging the gap between static policy learning and dynamic online replanning. However, existing MARL and Meta-RL approaches typically assume that task specifications are provided synchronously and remain stationary during execution, which is inconsistent with real-world multimedia systems, where information is naturally presented as asynchronous and heterogeneous streams (e.g., speech, visual cues, symbolic commands). This mismatch is particularly evident in human--robot interaction scenarios, where multimodal instructions arrive irregularly and are fundamentally misaligned with continuous control processes. From a multimedia perspective, this exposes a critical challenge: how to effectively align asynchronous multimodal inputs with continuous multi-agent decision-making. 

We formalize multimedia instruction following as a temporal alignment problem between irregular, semantically rich human inputs and high-frequency continuous control signals. Building upon the principles of Meta-MARL, we introduce VA-FastNavi-MARL, a novel framework for adaptive multi-arm motion planning and execution driven by heterogeneous visual and audio instructions. Unlike traditional MARL approaches that rely on fixed interaction strategies or static goal coordinates, VA-FastNavi-MARL learns from a dynamic distribution of multimodal human instructions. This capability enables the system to rapidly adapt its motion plans to unseen commands--ranging from voice directives to visual cues--with minimal gradient updates. Furthermore, we address the challenge of real-time responsiveness by implementing an asynchronous instruction stream, designed to handle the irregularity of human input. This facilitates seamless online control, allowing agents to incorporate dynamic, unprogrammed instructions into their execution pipeline instantaneously. Comprehensive experiments demonstrate that our method significantly outperforms MARL baselines in terms of both sample efficiency and real-time adaptation success.

\textbf{Our contributions include:}

\begin{itemize} 

\item A modality-agnostic multimedia instruction representation that aligns asynchronous audio, visual, and symbolic commands into a unified navigable task space, enabling instruction-level generalization.
\newpage
\item A real-time multimedia instruction stream formulation, modeling human inputs as an asynchronous temporal process and bridging bursty multimedia signals with continuous robotic control.

\item A Meta-RL based adaptive controller that leverages the above multimedia formulation for rapid policy adaptation in multi-robot systems.

\end{itemize}

\section{Related Works}

We categorize existing research into two streams: MARL for manipulation and Meta-RL for policy adaptation.

\subsection{MARL in Human-robot Interaction}
MARL offers a scalable alternative to centralized planners for high-dimensional human-robot interaction. Recent works have successfully applied MARL to task decomposition, such as separating position and orientation control \cite{10.1371/journal.pone.0311550} or employing modular strategies for surgical trajectory tracking \cite{10340943}. To address scalability in high-DOF systems, decentralized architectures have been proposed to learn local policies from global abstractions \cite{Shahid2021DecentralizedMC} or independent joint control \cite{Perrusquía2021}. Furthermore, innovations in reward shaping--ranging from Hindsight Experience Replay \cite{liu2021collaborative} to potential-based coordination \cite{yan2025markovpotentialgameconstruction}--have improved convergence. However, these methods typically overfit to fixed task definitions and struggle to generalize when semantic instructions or task sequences change. Our work bridges this gap by integrating a programmable instruction queue for real-time adaptation.

\subsection{Meta-Learning for Policy Adaptation}
Meta-RL enables agents to generalize to new tasks with minimal interaction. Optimization-based methods, pioneered by MAML \cite{MAML}, optimize initial parameters for rapid fine-tuning. Subsequent variants have improved stability through first-order approximations (e.g., Reptile, FOMAML \cite{FOMAML}) and implicit differentiation \cite{zhang2025directedmamlmetareinforcementlearning}. Alternatively, context-based methods like RL$^2$ \cite{rl2} and PEARL \cite{pearl} internalize adaptation by encoding past experience into latent states via RNNs or probabilistic encoders. Despite their success, standard Meta-RL formulations are primarily designed for single-agent settings. They lack the mechanisms to handle the non-stationarity of multi-agent coordination, particularly when agents must adapt to dynamic instruction streams rather than static physical environments.

\section{Method}

In this section, we introduce VA-FastNavi-MARL, a Meta-RL based motion planning framework that enables multi-arm robotic systems to rapidly adapt online to varying audio and visual instructions.

\subsection{Problem Formulation}

We consider a collaborative robotic motion control task where $N=3$ robotic arms must sequentially reach target points $\mathbf{p}_{\text{target}}$ derived from heterogeneous audio-visual instructions. We model the interaction among the arms as a \textit{Markov Game} (MG), defined by the tuple:
\[
(\mathcal{N}, \mathcal{S}, \mathcal{A}, \mathcal{P}, \mathcal{R}, \mathcal{I}, \gamma),
\]
where $\mathcal{N} = \{1, \dots, N\}$ denotes the set of agents. The instruction space $\mathcal{I}$ represents the set of latent embeddings derived from the multimedia buffer, where each instruction $\varphi \in \mathcal{I}$ specifies the target coordinates and execution sequence.

The joint state and action spaces are defined as $\mathcal{S} = \mathcal{S}_1 \times \cdots \times \mathcal{S}_N$ and $\mathcal{A} = \mathcal{A}_1 \times \cdots \times \mathcal{A}_N$, respectively, where $\mathcal{S}_i$ and $\mathcal{A}_i$ represent the individual state and action spaces of agent $i$. The state transition function is given by $\mathcal{P}: \mathcal{S} \times \mathcal{A} \rightarrow [0,1]$, and the reward function by $\mathcal{R}: \mathcal{S} \times \mathcal{A} \times \mathcal{I} \rightarrow \mathbb{R}$. We denote the discount factor by $\gamma \in (0,1]$.

To enable scalable coordination, we employ \textit{parameter sharing}, utilizing a unified policy $\pi_\theta$ across all agents. At time step $t$, the global observation $\mathbf{s}_t \in \mathbb{R}^{15}$ is constructed as a concatenation of individual agent states:
\[
\mathbf{s}_t = 
[\mathbf{p}^1_t, d^1_t, \phi^1_t;~
 \mathbf{p}^2_t, d^2_t, \phi^2_t;~
 \mathbf{p}^3_t, d^3_t, \phi^3_t],
\]
where $\mathbf{p}^i_t \in \mathbb{R}^3$ is the end-effector position of arm $i$, $d^i_t = \|\mathbf{p}^i_t - \mathbf{p}_{\text{target}}\|$ is the Euclidean distance to the target, and $\phi^i_t \in \{0,1,2\}$ is a discrete phase flag indicating the agent's current sequential status (e.g., waiting, executing, or completed). The joint action $\mathbf{a}_t \in \mathbb{R}^9$ concatenates the 3D velocity commands.

\subsection{Multimedia Instruction Generator}
\label{sec:input_module}

\begin{figure}[!h]
	\centering
	\includegraphics[width=0.87\linewidth]{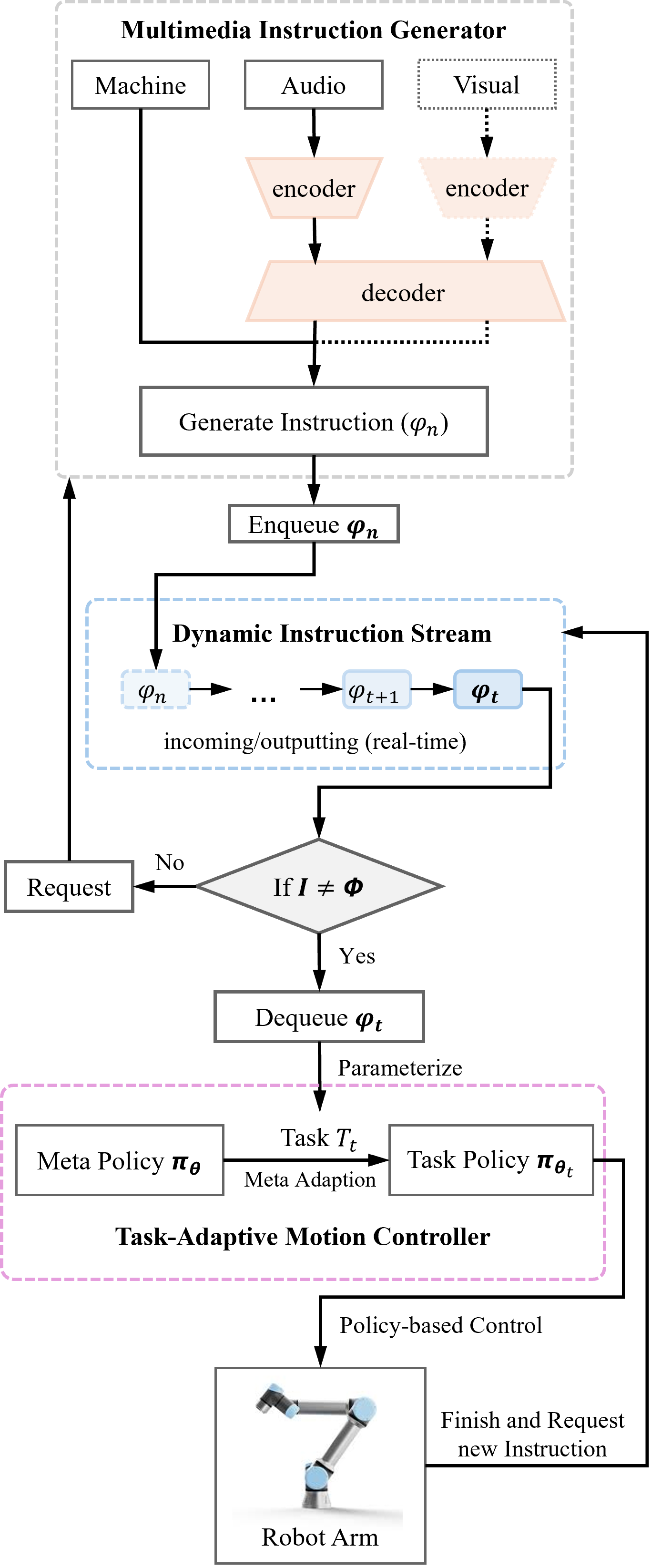}
	\caption{\textbf{Architecture of VA-FastNavi-MARL.} \textbf{Multimedia Instruction Generator (Top)} Heterogeneous inputs (machine, audio, visual) are fused into a unified embedding ($\varphi_n$) via parallel encoders. \textbf{ Dynamic Instruction Stream (Middle)} An asynchronous buffer schedules real-time commands ($\varphi_t$) to handle irregular input rates. \textbf{Task-adaptive Motion Controller (Bottom)} The active instruction parameterizes the task ($T_t$), triggering the Meta-RL policy $\pi_{\theta}$ to adapt into a specialized task policy $\pi_{\theta_t}$ for control.}
	\label{fig:pipeline}
\end{figure}

The proposed framework begins with a Multimedia-Driven Instruction Module (Fig.~\ref{fig:pipeline}, top) that unifies instructions from machine code ($\mathcal{I}_{machine}$), audio ($\mathcal{I}_{audio}$), and visual cues ($\mathcal{I}_{visual}$). To bridge the semantic gap between modalities, we employ a parallel Encoder-Decoder architecture. 

Rather than designing modality-specific feature extractors, we explicitly target semantic alignment across heterogeneous multimedia streams. To this end, we employ parallel encoder branches, using 1D convolutions to capture temporal phonetic patterns from $\mathcal{I}{audio}$ and 2D convolutions to extract spatial structures from $\mathcal{I}{visual}$, followed by a shared projection into a unified, modality-agnostic latent space $\Phi$. The key contribution of this module lies in cross-modal representation alignment, which bridges asynchronous and structurally distinct sensory inputs at the semantic level. By enforcing this alignment, the downstream policy $\pi_{\theta}$ becomes invariant to the source modality, treating audio and visual instructions as different manifestations of a shared navigable goal distribution. As a result, the agent can generalize instruction-following behaviors across modalities without requiring modality-specific policies.

\subsection{Dynamic Instruction Stream}

To enable continuous online supervision, we formulate the dynamic instruction stream (Fig.~\ref{fig:pipeline} Middle) as a Dynamic Instruction Queue (DIQ). At any time step $t$, the system maintains an ordered sequence of pending instructions $\mathcal{Q}_t$:

\begin{equation}
    \mathcal{Q}_t = (\varphi_1, \varphi_2, \dots, \varphi_K)
\end{equation}

where $\varphi_i$ represents the \textit{active instruction} currently guiding the control policy, and $\varphi_K$ is the last instruction in the buffer. The queue evolves according to a First-In-First-Out (FIFO) protocol governed by two primary transitions:

\textbf{Instruction Completion (Dequeue):} If the active instruction $\varphi_1$ is executed, the instruction will be removed, and control shifts to the subsequent objective:

\begin{equation}
    \mathcal{Q}_{t+1} \leftarrow \text{Dequeue}(\mathcal{Q}_t) = (\varphi_2, \dots, \varphi_K) \quad \text{if } s_t \models \varphi_1
\end{equation}
    
\textbf{Instruction Arrival (Enqueue):} New instructions $\varphi_{\text{new}}$ issued by the high-level planner or human operator are appended dynamically:

\begin{equation}
    \mathcal{Q}_{t+1} \leftarrow \text{Enqueue}(\mathcal{Q}_t, \varphi_{\text{new}}) = (\varphi_1, \dots, \varphi_K, \varphi_{\text{new}})
\end{equation}

Each motion instruction $\varphi_i$ is defined as an ordered sequence of waypoints, denoted as $\varphi_i = \langle w_{i,1}, \dots, w_{i,M} \rangle$, where $w_k$ represents the $k$-th sub-goal to be reached sequentially.

\subsection{VA-FastNavi-MARL}

\subsubsection{Task Distribution Formulation}
Standard MARL optimizes a policy for a single, static reward function. In contrast, VA-FastNavi-MARL requires the agent to generalize across a possibly infinite space of programmable motion instructions. We formulate this as a meta-learning problem where the task distribution $\rho(\mathcal{T})$ is defined over the distribution of valid instructions $\rho(\Phi)$.

A specific task $\mathcal{T}_i \sim \rho(\mathcal{T})$ is characterized by a unique instruction sequence $\varphi_i = \langle w_{i,1}, \dots, w_{i,M} \rangle$. The MDP for task $\mathcal{T}_i$ shares the same robot dynamics but possesses a task-specific reward function $R_i(s, a)$ that encourages sequential progression through the waypoints in $\varphi_i$. The goal of the meta-learner is to find a set of initial policy parameters $\theta$ that can rapidly adapt to maximize the expected return of any unseen instruction $\varphi_{\text{new}}$ after limited interaction.

\begin{algorithm}[h]
\caption{VA-FastNavi-MARL: Online Meta-Adaptation \& Execution}
\label{alg:online_adapt}
\begin{algorithmic}[1]
\State \textbf{Input:} Pre-trained Meta-Policy $\pi_\theta$, Adaptation step $\alpha$, Instruction Queue $\mathcal{Q}$
\State \textbf{Initialize:} Task Policy $\pi_{\theta_t} \leftarrow \pi_\theta$

\While{System Active}
    \State \textbf{1. Dynamic Instruction Retrieval}
    \If{$\mathcal{Q}$ is empty}
        \State Maintain \textit{Hold} state; \textbf{Continue}
    \Else
        \State $\varphi_t \leftarrow \mathcal{Q}.\text{dequeue}()$
    \EndIf

    \State \textbf{2. Fast Adaptation}
    \State Parameterize task $\mathcal{T}$ with $\varphi_t$
    \State Collect trajectories $\mathcal{D}_{\text{supp}}$ via interaction using $\pi_\theta$
    \State Compute task-specific parameters:
    \State $\quad \theta_t \leftarrow \theta - \alpha \nabla_\theta J_{\text{SAC}}(\mathcal{D}_{\text{supp}}, \theta)$
    \State \textbf{3. Task Execution}
    \While{Task $\varphi_t$ not completed}
        \State Select action $\mathbf{a}_t \sim \pi_{\theta_t}(\mathbf{s}_t, \varphi_t)$
        \State Execute $\mathbf{a}_t$, observe $\mathbf{s}_{t+1}$ and reward $r_t$
    \EndWhile
    \State Reset $\theta_t \leftarrow \theta$
\EndWhile
\end{algorithmic}
\end{algorithm}

\subsubsection{Meta-Training with Soft Actor-Critic (SAC)}
We employ MAML-based exploration using Soft Actor-Critic. The training process alternates between two loops:

\textbf{Inner Loop (Fast Adaptation):}
For a sampled instruction $\varphi_i$, the agent interacts with the environment to collect a \textit{support set} of transitions $\mathcal{D}_i^{\text{supp}}$. The policy parameters $\theta$ are updated to task-specific parameters $\theta'_i$ by performing $K$ gradient descent steps on the SAC objective $J_{\text{SAC}}$:
\begin{equation}
    \theta'_i = \theta - \alpha \nabla_\theta J_{\text{SAC}}(\theta, \mathcal{D}_i^{\text{supp}})
\end{equation}
This step simulates the agent's ability to "fine-tune" its understanding of the new instruction sequence geometry.

\textbf{Outer Loop (Meta-Optimization):}
To ensure the initialization $\theta$ is broadly generalizable, we evaluate the adapted parameters $\theta'_i$ on a separate \textit{query set} $\mathcal{D}_i^{\text{query}}$ (collected using $\pi_{\theta'_i}$). The meta-parameters are updated to minimize the loss across the batch of sampled instructions:
\begin{equation}
    \theta \leftarrow \theta - \beta \nabla_\theta \sum_{\mathcal{T}_i \sim p(\mathcal{T})} J_{\text{SAC}}(\theta'_i, \mathcal{D}_i^{\text{query}})
\end{equation}

\subsubsection{Task-adaptive Motion Controller (Fig.~\ref{fig:pipeline}, bottom)}
During Meta-Adaptation stage (Algorithm \ref{alg:online_adapt}), the robot receives novel, unseen instructions from  dynamic instruction stream. The system utilizes the meta-learned initialization $\theta$ and performs continuous online gradient updates. This allows the controller to dynamically adjust to the specific constraints of the current active instruction in the queue. We call these instruction adaptation steps as task-adaptive motion control. 

\section{Experiments}

Unlike prior works that focus on environment-level generalization, we evaluate generalization across instruction distributions. All experiments use the same workspace, while instruction semantics, length, modality, and temporal ordering differ between training and evaluation. 

\subsection{Experimental Setup}

\subsubsection{Environment Configuration}
\begin{figure}[h]
	\centering
	\includegraphics[width=0.45\linewidth]{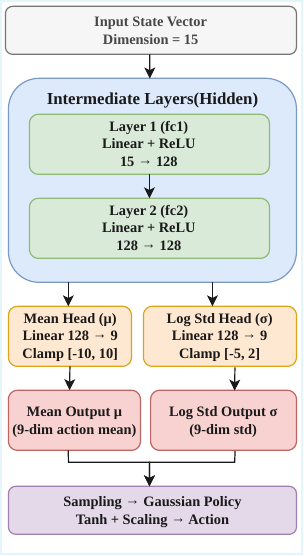}
	\caption{Architecture of the policy network. The network processes a 15-dimensional state vector through two hidden layers and uses a dual-head structure to output the parameters ($\mu$ and $\log \sigma$) of a 9-dimensional Gaussian policy. Output values are clamped for numerical stability before final action sampling and scaling.}
	\label{fig:model}
\end{figure}

We constructed a high-fidelity collaborative environment consisting of three robotic arms sharing a workspace. The state space $\mathcal{S} \in \mathbb{R}^{15}$ is defined by a unified feature vector containing the end-effector positions, distances to targets, and task-phase indicators for all three arms. The action space $\mathcal{A} \in \mathbb{R}^{9}$ consists of continuous velocity commands for the end-effectors. The architecture of the Policy network is shown in Fig. \ref{fig:model} 

\subsubsection{Robustness Evaluation Protocol} To validate the system's applicability to real-world scenarios where sensory inputs are often imperfect, we introduce stochastic perturbations during evaluation.

\textbf{Audio Noise:} We inject additive Gaussian white noise (SNR=20dB) into the raw audio waveforms to simulate environmental background chatter. 

\textbf{Visual Occlusion:} We apply random "Cutout" masks (covering 10-20\% of the image frame) to the visual inputs, simulating camera obstructions or sensor failures. The model is evaluated on its ability to maintain trajectory precision under these noisy conditions without re-training.

\subsubsection{Meta-Task Distribution}
In contrast to standard RL settings where the task is fixed, VA-FastNavi-MARL leverages a diverse distribution of tasks to facilitate meta-adaptation. We define a task $\mathcal{T}_i$ as a specific sequential coordination order of the three arms. During meta-training, the global policy is updated by sampling batches of $5$ distinct tasks. Specifically, the task distribution $p(\mathcal{T})$ is characterized by stochastic sequences of involving arbitrary arm ordering (e.g., $[2, 3, 1, 3, 2, 1]$), ensuring the agent learns to generalize across complex interaction patterns.

\subsection{Experimental Results}

We rigorously validate the performance of VA-FastNavi-MARL through evaluation along three core dimensions:

\begin{figure*}[h!] 
    \centering
    \subfigure{
        \includegraphics[width=0.30\linewidth]{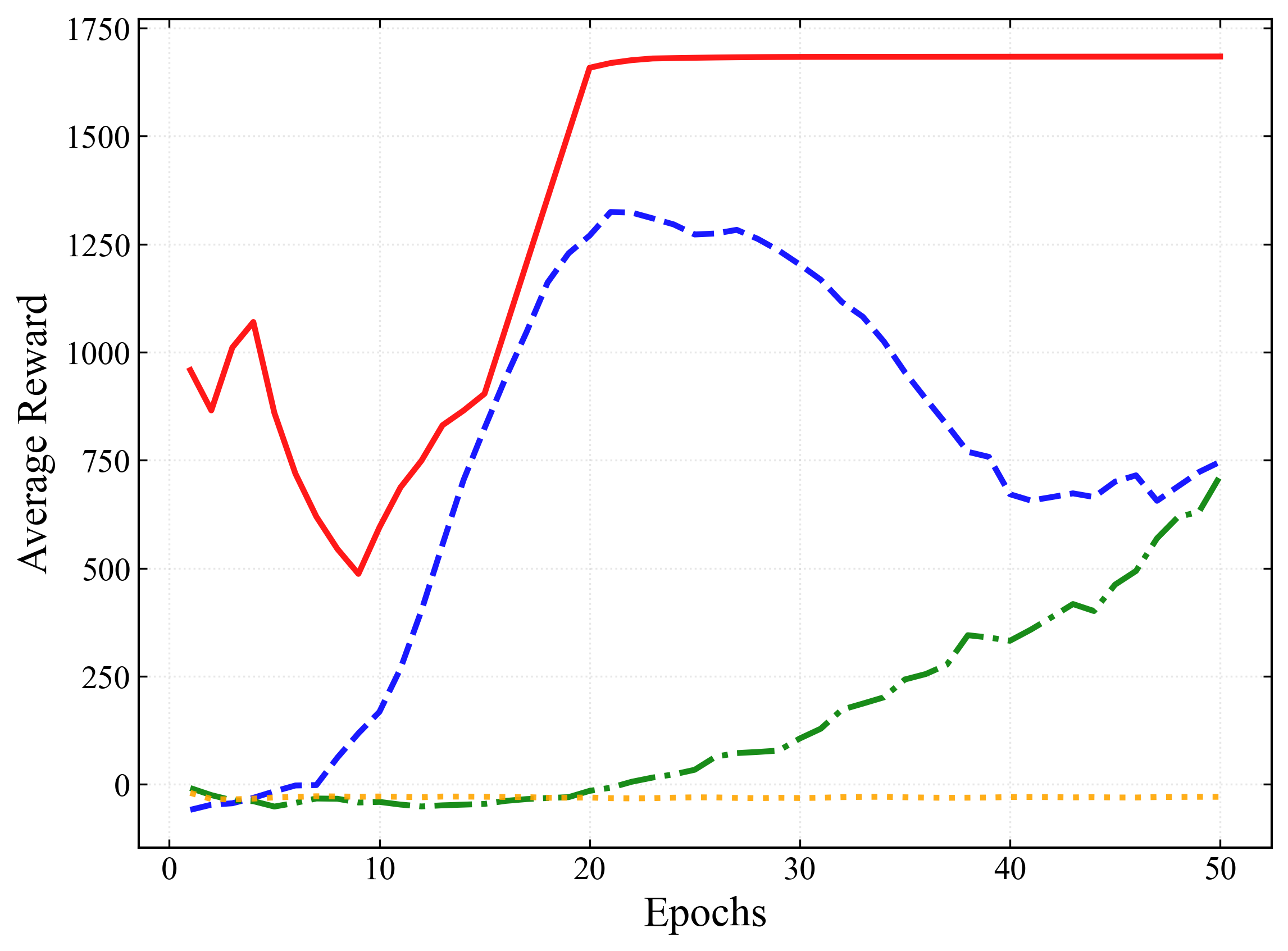}
        \label{fig:reward}    }
    \hfill
    \subfigure{
        \includegraphics[width=0.30\linewidth]{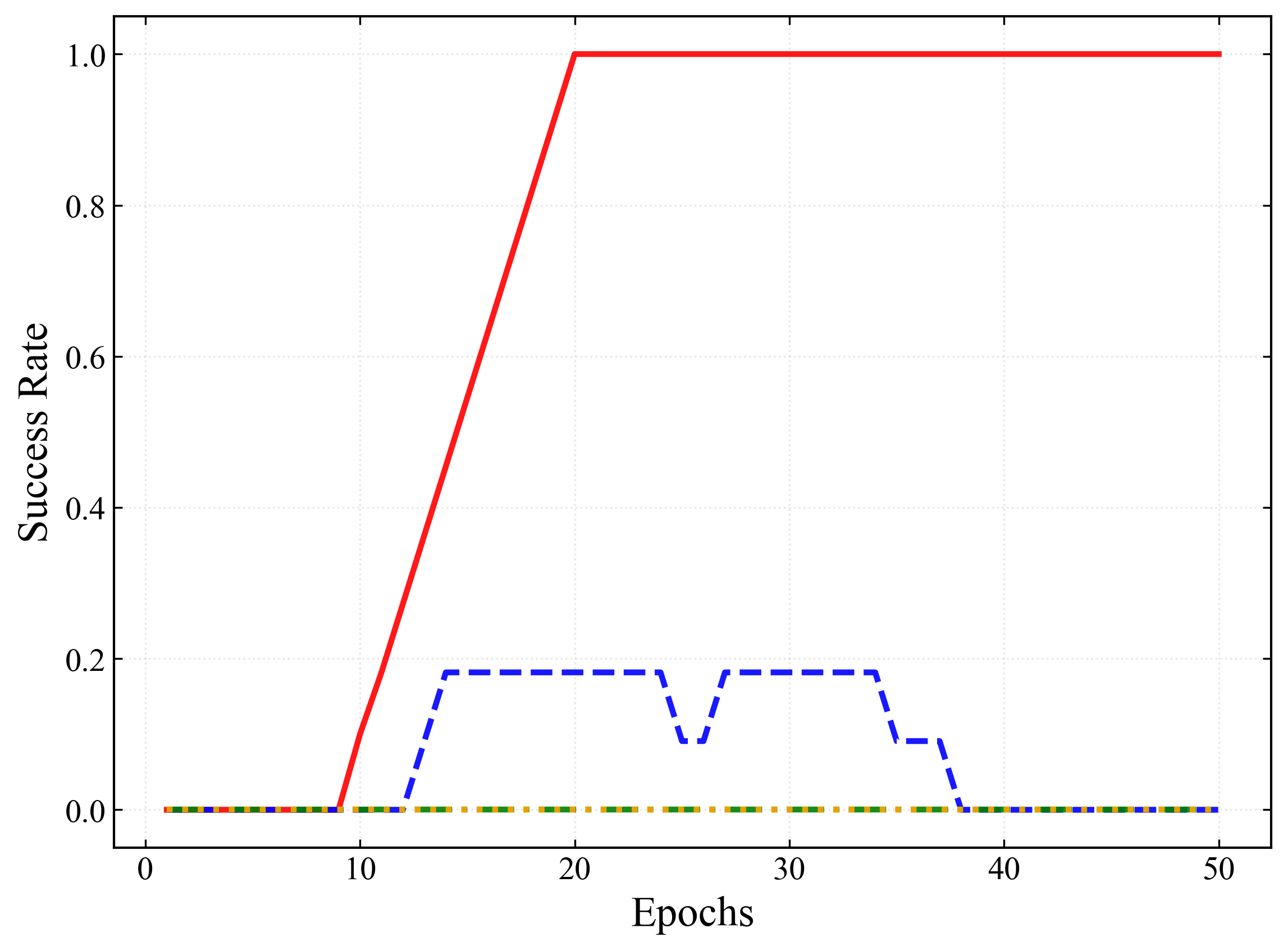}
        \label{fig:success}
    }
    \hfill 
    \subfigure{
        \includegraphics[width=0.30\linewidth]{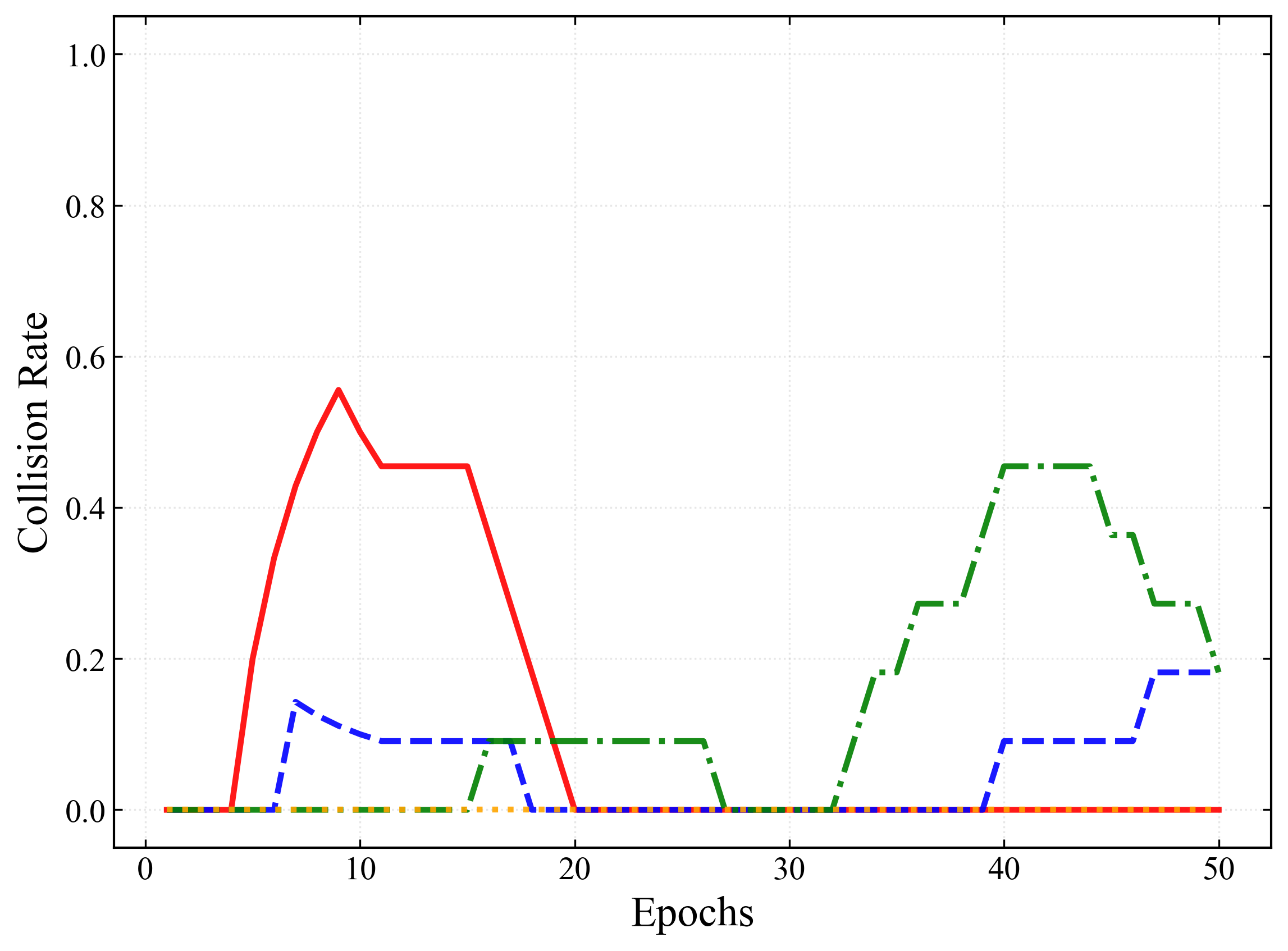}
        \label{fig:collision}
    }
    \vspace{-0.3cm} 
    \\ 
    \includegraphics[width=0.5\linewidth]{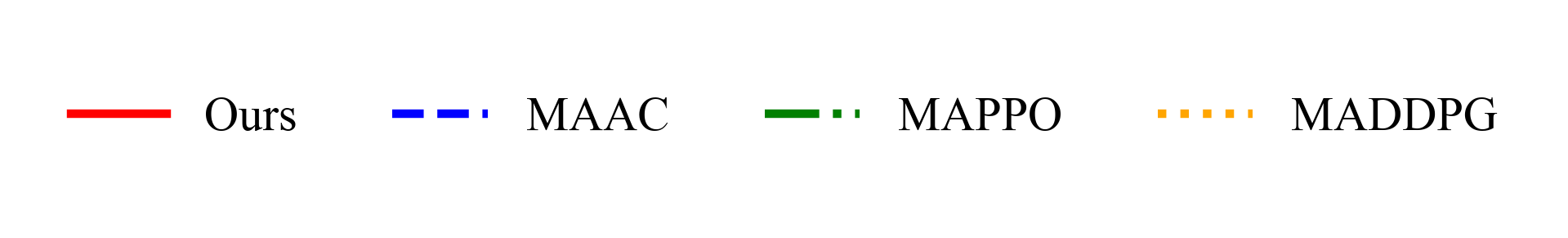}
    \caption{Performance comparison across different models. From left to right: Average Reward, Success Rate, and Collision Rate. Our method (Solid Red) achieves faster adaptation and higher safety compared to baselines.}
    \label{fig:comparison}
\end{figure*}

\begin{figure*}[h!]
    \centering
    \includegraphics[width=0.95\linewidth]{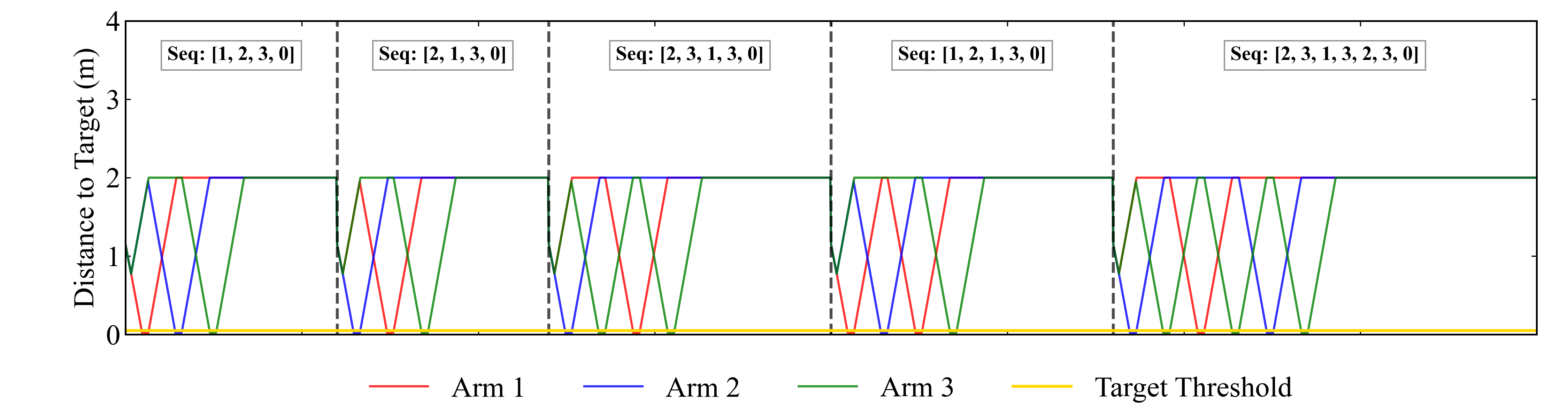}

    \includegraphics[width=0.95\linewidth]{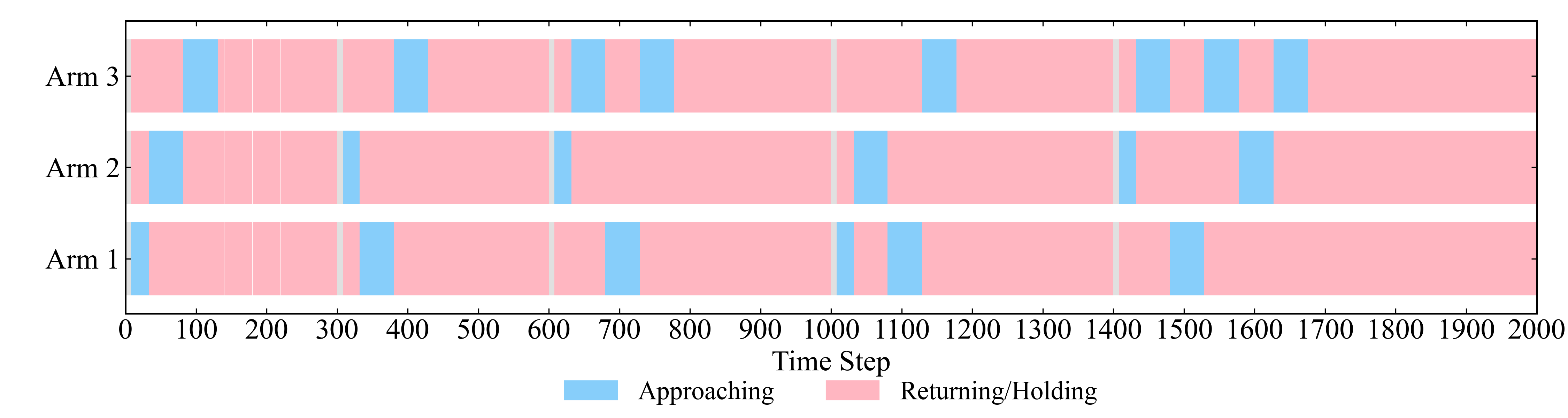}
    
    \caption{Trajectory evolution (Top) during the Long-Horizon Continuous Adaptation test ($T=2000$ steps).The experiment is segmented into five distinct phases (separated by vertical dashed lines), where the underlying instruction logic shifts abruptly from simple instructions (e.g., \texttt{Seq: [1,2,3]}) to complex instructions (e.g., \texttt{Seq: [2,3,1,3,2,3]}). Bottom is the state activation timeline corresponding to the continuous adaptation task. Blue segments denote the \textit{Approaching} phase (active execution), while pink regions represent the \textit{Returning/Holding} phase.} 
    \label{fig:long_horizon_distance}

\end{figure*}

\subsubsection{Convergence Speed} The provided performance metrics (As shown in Fig.~\ref{fig:comparison}) illustrate a comparative analysis between our proposed method and three baselines (MAAC, MAPPO, and MADDPG) across 50 adaption steps. Our method (solid red line) demonstrates superior convergence, reaching a peak Average Reward of approximately 1700 by epoch 20, whereas baselines like MAAC suffer from significant performance degradation after an initial rise. This efficiency is mirrored in the Success Rate, where our approach achieves a perfect 1.0 score early on, while MAPPO and MADDPG fail to solve the task. Furthermore, despite an initial exploration-driven spike in the Collision Rate, our method successfully converges to zero collisions by epoch 20. These results highlight our model's robustness and its ability to achieve optimal multi-agent coordination while prioritizing safety and task completion.

\subsubsection{Robot Control Performance}
Beyond single-task performance, a critical requirement for real-world deployment is the system's ability to handle dynamic, non-stationary instruction streams, which we visualized in the Long-Horizon Continuous Adaptation Test (Fig.~\ref{fig:long_horizon_distance}). The experiment tracks the distance-to-target trajectories of the three arms over 2000 steps, divided into five phases separated by vertical dashed lines; the underlying task logic is abruptly switched when previous instruction has finished, evolving from a simple (\texttt{Seq:[1,2,3]}) to more irregular patterns such as \texttt{Seq:[2,3,1,3]}. The plotted trajectories reveal a robust ``negotiated turn-taking'' behaviour: for instance, during the third phase (steps 600--1000) that demands the four-step sequence  \texttt{[2,3,1,3]}, the system modulates the activity of Arm 2 (blue) and Arm 3 (green) to accommodate the interleaved activation of Arm 1 (red), with distance valleys aligning precisely to the prescribed order. Immediately after each phase boundary, the agents adjust their policy within a few gradient steps--seen as rapid corrections in trajectory slope--without catastrophic forgetting or collisions, and the yellow threshold line confirms that every targeted arm converges to within the tolerance  $\epsilon_{\text{target}}$ for every instruction in the queue.

\subsubsection{Efficiency and Robustness Analysis on Instruction Generation} In this part, we evaluate the performance of instruction generation of different multimedia input (machine order, audio input and audio-visual input) in terms of efficiency (instruction generation time) and accuracy (instruction generation precision). To investigate the robustness of VA-FastNavi-MARL under instructions of varying complexity, we categorize instructions into three difficulty levels based on their semantic and compositional complexity, which is approximated by instruction length. Specifically, instructions are labeled as \textbf{Easy} ($[0,3]$), \textbf{Medium} ($[4,6]$), and \textbf{Hard} ($[7,10]$). This evaluation protocol mirrors standard multimedia generalization benchmarks, where the sensory stream remains similar while the semantic instruction distribution shifts.

\begin{table}[h]
\centering

\caption{Instruction Generation Time and Accuracy of VA-FastNavi-MARL Across Instruction Difficulty Levels and Multimedia Inputs (5 Seeds, 100 Samples)}
\begin{tabular}{clll}
\hline
\multicolumn{1}{l}{Difficulty} & Input Type        &  Time (s) & Accuracy (\%) \\ \hline
\multirow{3}{*}{Easy}                & Audio        &             0.008$\pm$0.001         &     93.80$\pm$1.70   \\
                                     & Audio (noise)      &   0.008$\pm$0.001         &     87.40$\pm$2.70        \\ 
                                     & Audio-Visual &0.389$\pm$0.038             & 94.60$\pm$0.80   \\ 
                                     & Audio-Visual (noise) &  0.498$\pm$0.058            & 92.80$\pm$2.14    \\ 
                                     \hline
\multirow{3}{*}{Medium}              & Audio        &             0.015$\pm$0.001       &     87.80$\pm$1.60  \\
                                     & Audio (noise)      &        0.017$\pm$0.001     & 77.20$\pm$2.00            \\ 
                                     & Audio-Visual &   0.465$\pm$0.047 &          91.20$\pm$2.04      \\ 
                                     & Audio-Visual (noise) &   0.571$\pm$0.034         &   88.00$\pm$3.74 \\ 
                                     \hline 
                                     
\multirow{3}{*}{Hard}                & Audio        &             0.023$\pm$0.002       &     83.00$\pm$1.50  \\
                                     & Audio (noise)      &     0.025$\pm$0.003        &  67.00$\pm$3.50           \\ 
                                     & Audio-Visual &   0.489$\pm$0.028 &          90.60$\pm$2.24     \\ 
                                     & Audio-Visual (noise) &   0.516$\pm$0.079          &   83.20$\pm$1.60 \\ 
                                     \hline 

\end{tabular}
\label{tab:time}
\end{table}

Table \ref{tab:time} validates the system's efficiency and resilience across varying difficulty levels. The generation time remains strictly sub-second ($<0.6s$) even under noisy conditions, fully satisfying real-time control requirements. Crucially, the system demonstrates remarkable robustness against sensory perturbations. In \textbf{Easy} and \textbf{Medium} settings, the impact of noise is minimal, with the model consistently maintaining high precision. This stability extends to the challenging \textbf{Hard} category, where our Audio-Visual fusion significantly outperforms the unimodal baseline (retaining $83.20\%$ accuracy versus $67.00\%$), confirming that our modality-agnostic embedding effectively preserves robust semantic intent against environmental corruption.

\section{Conclusion}

In this paper, we introduced VA-FastNavi-MARL, a Meta-RL framework designed for time-critical multi-arm adaptation to dynamic motion instructions. Our Multimedia-Driven Instruction Generator utilizes parallel encoder-decoder networks to fuse heterogeneous inputs--machine codes, audio, and visual cues--into a unified, modality-agnostic instruction. Crucially, by leveraging an asynchronous programmable motion-instruction stream, the framework decouples complex perception processing from actuation. This design supports highly responsive human-robot interaction, allowing users to issue online motion commands that the system can interpret and execute with millisecond latency. The demonstrated computational efficiency and robustness to sensory noise confirm the system's potential for deployment on physical robotic hardware. Future work will focus on sim-to-real transfer using physical UR5 robotic arms to validate robustness against real-world communication delays, as well as extending the framework to larger robot teams.

\bibliographystyle{IEEEbib}
\bibliography{icme2026references}

\end{document}